\documentclass{article}

\usepackage[preprint]{neurips_2026}
\usepackage[utf8]{inputenc}
\usepackage[T1]{fontenc}
\usepackage{amsmath}
\usepackage{amssymb}
\usepackage{booktabs}
\usepackage{graphicx}
\usepackage{microtype}
\usepackage{xcolor}
\usepackage{url}
\usepackage{algorithm}
\usepackage{algpseudocode}

\title{Liquid Latent State Dynamics for Interpretable Turbofan Degradation Modeling}

\author{
  Weizhi Nie, Weijie Wang, Yuting Su \\
  Tianjin University
}

\begin{document}

\maketitle

\begin{abstract}
Multivariate time-series models for prognostics are often evaluated by point prediction accuracy, yet their internal states rarely expose a coherent degradation process. We study liquid neural networks as latent dynamics models for aircraft engine health monitoring on the C-MAPSS benchmark. The proposed model encodes a history window into a latent state, evolves that state with a liquid transition model, and decodes future sensor observations. To separate health evolution from operating-condition variation, the latent state is factorized into degradation and condition components. Remaining useful life, monotonic risk, and latent-consistency losses supervise the degradation component, while condition prediction and decorrelation losses discourage operating-condition leakage. Across FD001--FD004, the full disentangled model improves overall sensor forecasting RMSE from 0.2438 for a GRU baseline to 0.2266, with the largest gains on the multi-condition subsets FD002 and FD004. The learned degradation state also forms a clearer temporal degradation axis, reaching an average state-speed Spearman correlation of 0.5960. Direct remaining-useful-life regression remains stronger for the GRU baseline, indicating that the proposed representation is currently more effective as an interpretable world model for degradation dynamics than as a calibrated lifetime regressor. These results suggest that liquid latent dynamics can bridge predictive maintenance forecasting and inspectable health-state modeling.
\end{abstract}

\section{Introduction}

Reliable prognostics for complex engineering systems requires more than accurate
one-step prediction. In aircraft engine monitoring, the measured signal is a
multivariate time series of sensor readings and operating settings, while the
quantity of interest is an unobserved health state that evolves over time. A
model can therefore obtain low forecasting error by exploiting short-term
sensor regularities without learning a state representation that corresponds to
degradation. This distinction matters in predictive maintenance: practitioners
need not only a forecast of future measurements or remaining useful life (RUL),
but also an inspectable trajectory that explains how the system is moving toward
failure.

The C-MAPSS turbofan benchmark makes this issue especially clear. Its subsets
contain different combinations of fault modes and operating conditions, so
changes in sensor values may reflect either health deterioration or benign
condition shifts. Standard recurrent forecasting models can absorb both effects
into a single hidden state. Such a representation may be sufficient for direct
regression, but it is difficult to interpret as an engine health state, and it
does not explicitly define how the state should be rolled forward as a dynamical
system. A central challenge is therefore to learn a latent state that is useful
for forecasting while preserving a meaningful separation between degradation
dynamics and operating-condition variation.

\begin{figure}[t]
  \centering
  \includegraphics[width=\linewidth]{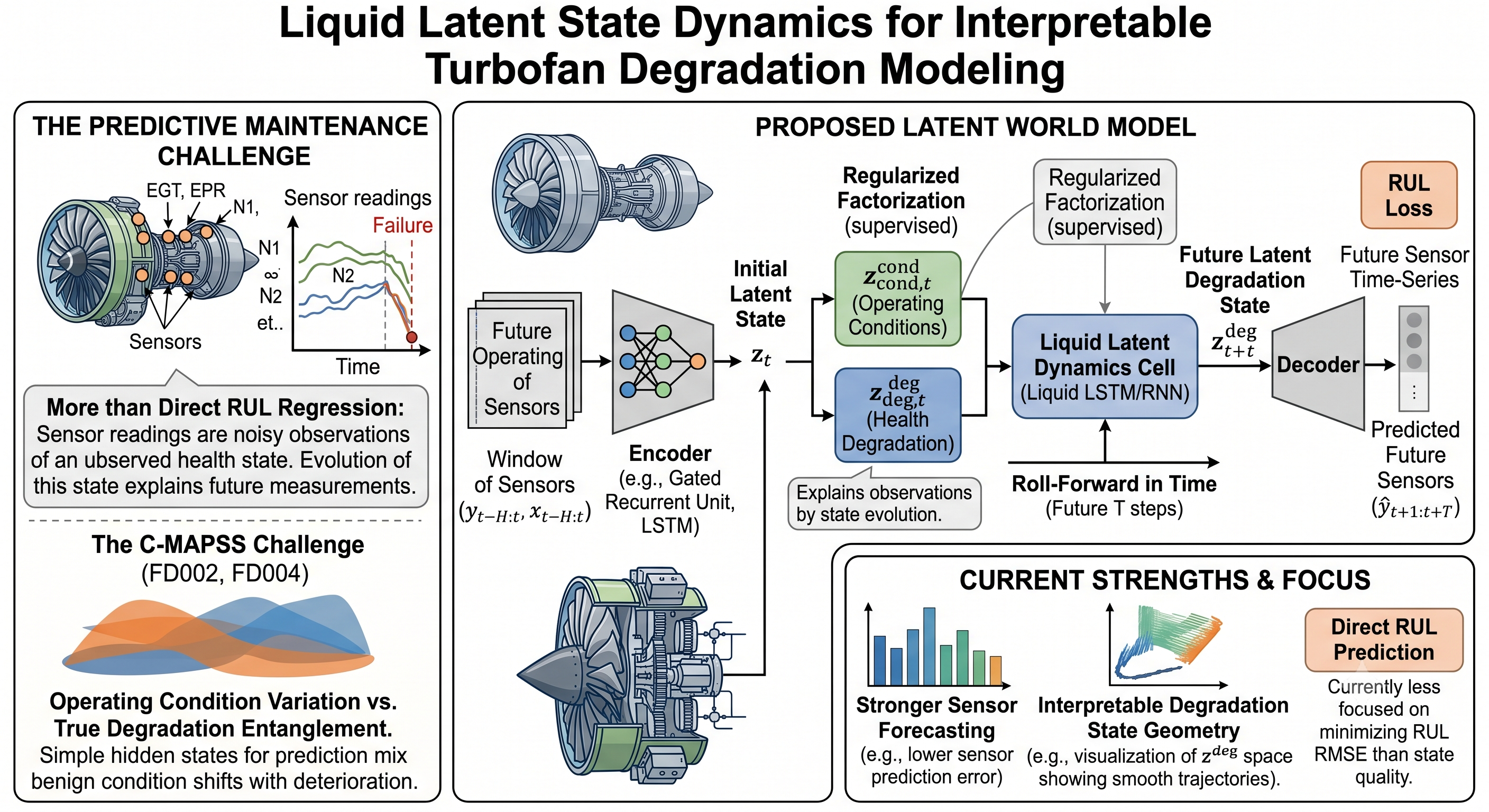}
  \caption{Motivation for learning latent degradation dynamics. Sensor
  observations mix health degradation with operating-condition variation; the
  desired representation is an interpretable hidden-state trajectory whose
  evolution exposes the degradation trend behind future observations.}
  \label{fig:motivation}
\end{figure}

We study liquid neural networks as latent dynamics models for this
setting. Given a history window, an encoder maps the observed sequence to a
latent state. A liquid transition model then evolves the state forward, and a
decoder reconstructs future sensor observations. This design treats forecasting
as the consequence of latent state evolution rather than as direct sequence
extrapolation. To reduce condition-induced entanglement, the latent state is
factorized into a degradation component and a condition component. RUL,
monotonic risk, and latent-consistency losses are applied to the degradation
component, while condition prediction and decorrelation losses encourage the
condition component to capture operating context.

The resulting empirical picture is mixed but informative. On C-MAPSS
FD001--FD004, the proposed disentangled liquid latent model improves overall
sensor forecasting RMSE from 0.2438 for a GRU baseline to 0.2266, with the
largest gains on FD002 and FD004, the subsets with more complex operating
conditions. The learned degradation component also exhibits a clearer temporal
degradation axis, with an average state-speed Spearman correlation of 0.5960.
However, the method does not yet outperform the GRU baseline on direct RUL RMSE.
This limitation is important: the current model is best understood as an
interpretable latent world model for degradation dynamics, rather than as a
fully calibrated lifetime regressor.

This paper makes the following contributions:
\begin{itemize}
  \item A liquid latent dynamics formulation for multivariate turbofan sensor
  forecasting, in which future observations are generated by rolling forward an
  explicit latent state.
  \item A disentangled state design that separates degradation and operating
  condition factors and restricts RUL-oriented supervision to the degradation
  component.
  \item An empirical evaluation on all four C-MAPSS subsets showing improved
  sensor forecasting under complex operating conditions and more coherent
  degradation-state geometry, together with a transparent analysis of the
  remaining RUL calibration gap.
\end{itemize}

\section{Related Work}

\paragraph{Remaining useful life estimation.}
The C-MAPSS turbofan benchmark is a standard testbed for data-driven
prognostics because it provides run-to-failure trajectories under different
fault and operating-condition settings \citep{saxena2008damage}. More broadly,
prognostics and health management has long studied the estimation of degradation
state and remaining useful life from condition-monitoring signals
\citep{schwabacher2005survey,jardine2006review,si2011review,lei2018review,zhao2017health,gao2020survey}.
A large body of work
frames C-MAPSS primarily as a supervised RUL regression problem. Early neural
approaches already used recurrent models for engine RUL estimation
\citep{heimes2008rnn}, while later work adopted LSTMs and sequence embeddings
to learn temporal features directly from sensor windows
\citep{zheng2017lstm,gugulothu2017embed,malhotra2016lstm}. Convolutional and
temporal-convolutional variants improve local temporal feature extraction
\citep{babu2016dcnn,li2018dcnn,jayasinghe2019tcmn}, and attention-based or
Transformer-style architectures provide more flexible feature aggregation
\citep{vaswani2017attention,lim2021tft,zhou2021informer}. These models are
strong baselines for direct lifetime regression, and our experiments confirm
that a GRU remains competitive for RUL RMSE. The goal of this paper is
therefore not to claim a new RUL-only state of the art. Instead, it asks
whether a prognostic model can learn an explicit latent degradation dynamics
that remains useful for forecasting and inspection.

\paragraph{Latent dynamics and world models.}
Learning a compact state whose evolution explains future observations is a
central idea in latent dynamics models and world models. Variational recurrent
and state-space models combine latent variables with learned transition models
for sequential data \citep{chung2015vrnn,krishnan2015dvbf,fraccaro2016dkf}.
In model-based reinforcement learning, world models learn compressed
representations and predict their future evolution to support planning or
policy learning \citep{ha2018worldmodels,hafner2019planet}. This perspective is
attractive for prognostics because sensor readings are observations of an
underlying physical process rather than the process itself. However, most
industrial RUL models optimize the final prediction target directly, leaving
the learned hidden state underconstrained. Our work adopts the latent dynamics
view but applies it to predictive maintenance: the latent state is rolled
forward to forecast future sensors, and its geometry is evaluated as a
candidate degradation state.

\paragraph{Continuous-time neural dynamics.}
Neural ordinary differential equations introduced a general framework for
learning continuous transformations by parameterizing the derivative of a hidden
state with a neural network \citep{chen2018neuralode}. Subsequent work extended
this idea to irregularly sampled time series through ODE-RNNs and Latent ODEs
\citep{rubanova2019latentode}. Liquid time-constant networks belong to the same
broad family of continuous-time recurrent models, but use input- and
state-dependent time constants to obtain adaptive hidden dynamics
\citep{hasani2021ltc}. Related continuous-time recurrent architectures include
neural circuit policies and closed-form continuous-depth models
\citep{lechner2020ncp,hasani2022cfc}. This property is appealing for
degradation modeling, where the rate of change may vary with both health state
and operating condition. Existing liquid-network work has mainly emphasized
expressivity, stability, and sequence prediction. We use liquid dynamics more
narrowly: as the transition operator of a prognostic latent state whose
degradation-related structure can be inspected.

\paragraph{Disentanglement under operating conditions.}
A persistent difficulty in turbofan prognostics is that sensor variation mixes
two factors: degradation and operating condition. Multi-condition C-MAPSS
subsets such as FD002 and FD004 are therefore harder than single-condition
subsets because a model may mistake benign operating shifts for health changes,
or hide both factors in the same representation. Prior RUL models often address
this issue implicitly through normalization, windowing, attention, or larger
sequence encoders. These choices can improve predictive accuracy, but they do
not by themselves specify which part of the representation corresponds to
degradation. Representation-learning work has shown that disentanglement can be
encouraged through architectural and objective-level constraints, but also that
unsupervised disentanglement is fragile without suitable inductive bias
\citep{bengio2013representation,higgins2017betavae,locatello2019challenging}.
Our approach therefore makes the intended separation explicit by factorizing the
latent state into degradation and condition components, supervising RUL and
monotonic risk only through the degradation component, and penalizing
correlation with operating settings. The resulting separation is imperfect, as
condition states still contain some degradation information, but it provides a
concrete mechanism for testing whether a learned prognostic state is
interpretable rather than merely predictive.

\section{Method}

\begin{figure}[t]
  \centering
  \includegraphics[width=\linewidth]{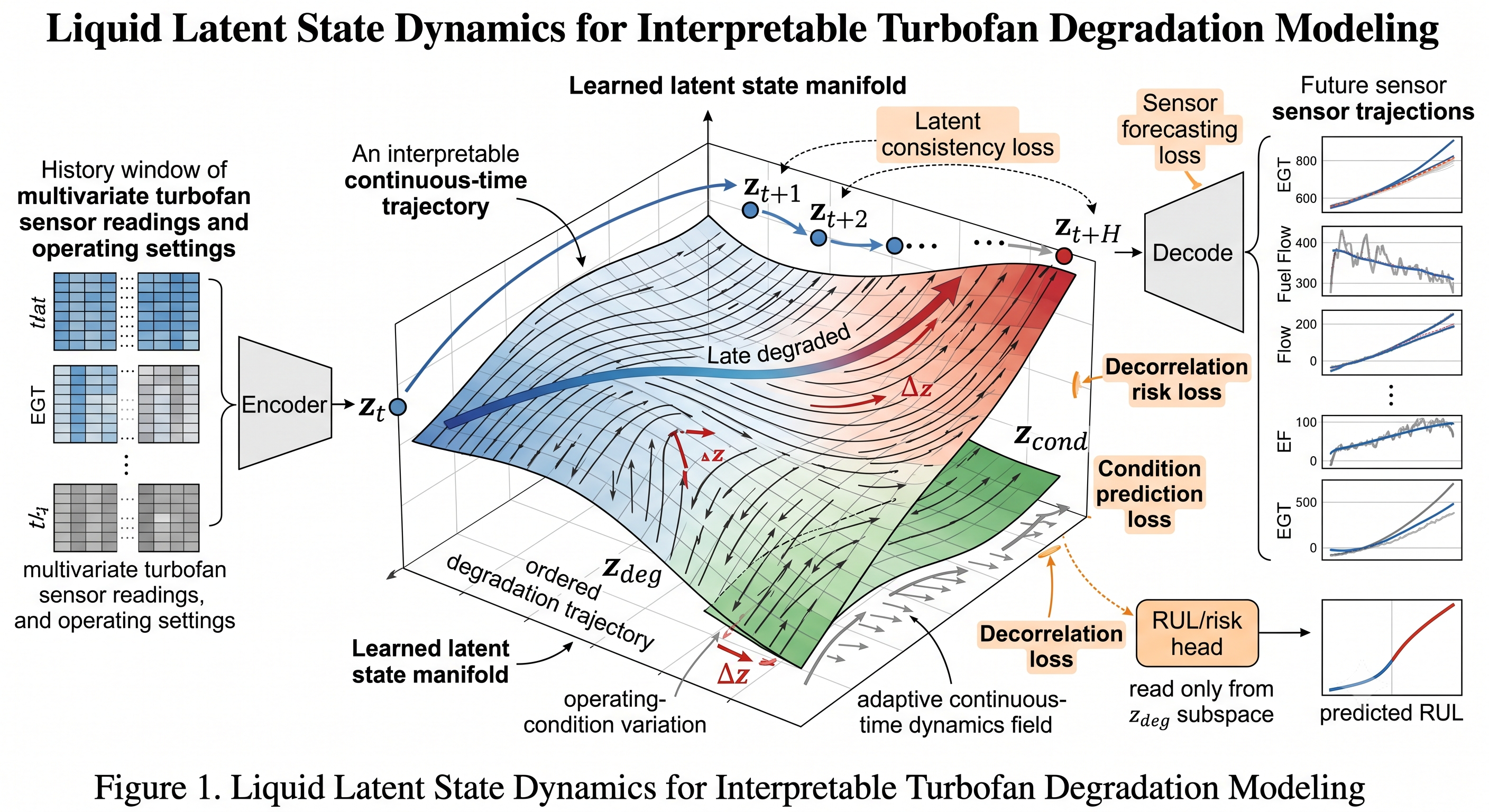}
  \caption{Framework of the proposed liquid latent dynamics model. The method
  encodes a history window into degradation and condition states, rolls the
  degradation state forward with liquid dynamics conditioned on future operating
  settings, and decodes the resulting hidden-state trajectory into future sensor
  predictions. The central object is the learned evolution trend of the latent
  degradation state, rather than a direct sequence-to-sequence mapping.}
  \label{fig:framework}
\end{figure}

\subsection{Problem Formulation}

Let \(x_t \in \mathbb{R}^{d_x}\) denote the observed sensor vector at cycle
\(t\), and let \(a_t \in \mathbb{R}^{d_a}\) denote the corresponding operating
setting. In C-MAPSS, \(d_a=3\) and \(d_x=21\). We write
\(u_t=[a_t,x_t]\in\mathbb{R}^{d_a+d_x}\) for the full observation at one cycle.
Given a history window of length \(L\),
\begin{equation}
  X_t = [u_{t-L+1}, \ldots, u_t], \qquad
  u_t = [a_t, x_t],
\end{equation}
the model predicts future sensor observations \(\hat{x}_{t+1:t+H}\) over a
horizon \(H\), given future operating settings \(a_{t+1:t+H}\). It also
estimates a normalized RUL target at the end of the horizon. We use
\(z_t\in\mathbb{R}^{d_z}\) to denote the learned latent system state, and
\(\Delta z_{t+h}=z_{t+h}-z_{t+h-1}\) to denote the learned state increment
during one rollout step. The central modeling assumption is that the sensor
readings are observations of an underlying state, rather than the state itself:
\begin{equation}
  x_t \sim p(x_t \mid z_t, a_t).
\end{equation}
The objective is therefore to learn a state representation whose rollout
explains future observations and whose state increments reflect degradation
progress.

\subsection{Liquid Latent Dynamics}

The basic model consists of three components: an encoder, a liquid transition
operator, and a decoder. A recurrent encoder maps the history window to an
initial latent state,
\begin{equation}
  z_t = E_\theta(X_t).
\end{equation}
The latent state is then rolled forward with a setting-conditioned liquid cell,
\begin{equation}
  z_{t+h}, \Delta z_{t+h}
  = G_\psi(z_{t+h-1}, a_{t+h}), \qquad h=1,\ldots,H.
\end{equation}
In implementation, the liquid cell computes a target drift and an adaptive time
constant. For compactness, define
\(c_{t+h}=[z_{t+h-1},a_{t+h}]\) as the transition input. The cell computes
\begin{align}
  m_{t+h} &= f_\psi(c_{t+h}), \\
  \tau_{t+h} &= \operatorname{softplus}(g_\psi(c_{t+h})) + \epsilon, \\
  \gamma_{t+h} &= 1 - \exp(-\Delta t / \tau_{t+h}), \\
  \Delta z_{t+h} &= \gamma_{t+h} \odot (m_{t+h} - z_{t+h-1}), \\
  z_{t+h} &= z_{t+h-1} + \Delta z_{t+h}.
\end{align}
Here \(f_\psi\) and \(g_\psi\) are neural networks, \(m_{t+h}\) is the
state-dependent drift target, \(\tau_{t+h}\) is an adaptive time constant,
\(\epsilon\) prevents numerical degeneracy, \(\gamma_{t+h}\in(0,1)\) is the
learned update gate, and \(\odot\) denotes elementwise multiplication. The key
quantity is \(\Delta z_{t+h}\): the model does not predict an unconstrained
next state directly. Instead, it learns how far the current state should move
toward a locally predicted drift target, with the step size controlled by the
time constant. A small \(\tau\) produces a larger gate and faster movement,
whereas a large \(\tau\) produces a smaller gate and slower movement. The
liquid transition therefore parameterizes the trend of hidden-state change.

The state increment is learned indirectly through several constraints. Sensor
forecasting gradients force the rolled-out states to reconstruct
\(x_{t+1:t+H}\). Latent consistency pulls the rollout state toward the state
obtained by re-encoding the observed future window. The monotonic risk loss
biases the degradation direction so that risk does not decrease along the
rollout. Finally, the smoothness penalty controls the magnitude of
\(\Delta z\), preventing the latent trajectory from becoming an arbitrary jump
process. This is a discrete rollout of adaptive continuous-time recurrent
dynamics, but we treat it as a learned transition model rather than an
identified physical law. Future sensor values are
decoded from the rolled-out states:
\begin{equation}
  \hat{x}_{t+h} = D_\phi(z_{t+h}).
\end{equation}
A risk/RUL head also reads from the final latent state,
\begin{equation}
  \widehat{\mathrm{RUL}}_{t+H} = Q_\eta(z_{t+H}).
\end{equation}

\subsection{Disentangled Degradation and Condition State}

The basic latent state can mix degradation and operating condition. To make the
intended structure explicit, the proposed model factorizes the encoded state as
\begin{equation}
  z_t = [z_t^{\mathrm{deg}}, z_t^{\mathrm{cond}}],
\end{equation}
where \(z_t^{\mathrm{deg}}\) is intended to carry health and degradation
information, and \(z_t^{\mathrm{cond}}\) is intended to carry operating-context
information. The encoder produces the two components through separate heads:
\begin{equation}
  z_t^{\mathrm{deg}}, z_t^{\mathrm{cond}} = E_\theta(X_t).
\end{equation}
During rollout, the condition state is updated from the previous condition
state and the future operating setting,
\begin{equation}
  z_{t+h}^{\mathrm{cond}}
  = C_\omega(z_{t+h-1}^{\mathrm{cond}}, a_{t+h}),
\end{equation}
while the degradation state is evolved by a liquid transition conditioned on
both the operating setting and the current condition state:
\begin{equation}
  z_{t+h}^{\mathrm{deg}}, \Delta z_{t+h}^{\mathrm{deg}}
  =
  G_\psi\!\left(
    z_{t+h-1}^{\mathrm{deg}},
    [a_{t+h}, z_{t+h}^{\mathrm{cond}}]
  \right).
\end{equation}
In this form, the learned degradation increment
\(\Delta z_{t+h}^{\mathrm{deg}}\) is the main object of interest. It describes
how the health-related state changes from one cycle to the next after
accounting for operating condition. The condition state does not directly
determine RUL; instead, it modulates the degradation transition by indicating
the operating regime. This design prevents the model from explaining all sensor
variation with a single hidden vector and allows us to test whether the
degradation subspace follows a coherent trajectory.
The decoder uses both factors,
\begin{equation}
  \hat{x}_{t+h}
  =
  D_\phi(z_{t+h}^{\mathrm{deg}}, z_{t+h}^{\mathrm{cond}}),
\end{equation}
but the prognostic head uses only the degradation component:
\begin{equation}
  \widehat{\mathrm{RUL}}_{t+H}
  =
  Q_\eta(z_{t+H}^{\mathrm{deg}}).
\end{equation}
We define a normalized risk score as
\begin{equation}
  r_{t+h} = 1 - Q_\eta(z_{t+h}^{\mathrm{deg}}),
\end{equation}
so that increasing risk corresponds to decreasing predicted remaining life.

\subsection{Training Objective}

The model is trained with a multi-term objective that combines forecasting
accuracy, RUL supervision, latent rollout consistency, monotonic risk, condition
prediction, decorrelation, and a smoothness penalty:
\begin{align}
  \label{eq:total-loss}
  \mathcal{L}
  &=
  \mathcal{L}_{\mathrm{sensor}}
  + \lambda_{\mathrm{rul}}\mathcal{L}_{\mathrm{RUL}}
  + \lambda_{\mathrm{latent}}\mathcal{L}_{\mathrm{latent}}
  + \lambda_{\mathrm{mono}}\mathcal{L}_{\mathrm{mono}} \nonumber \\
  &\quad
  + \lambda_{\mathrm{cond}}\mathcal{L}_{\mathrm{cond}}
  + \lambda_{\mathrm{decor}}\mathcal{L}_{\mathrm{decor}}
  + \lambda_{\mathrm{smooth}}\mathcal{L}_{\mathrm{smooth}}.
\end{align}
The sensor forecasting loss is
\begin{equation}
  \mathcal{L}_{\mathrm{sensor}}
  =
  \frac{1}{H}\sum_{h=1}^{H}
  \left\|
    \hat{x}_{t+h} - x_{t+h}
  \right\|_2^2.
\end{equation}
The RUL loss supervises the normalized RUL prediction at the end of the
forecasting horizon:
\begin{equation}
  \mathcal{L}_{\mathrm{RUL}}
  =
  \left\|
    \widehat{\mathrm{RUL}}_{t+H} - \mathrm{RUL}_{t+H}
  \right\|_2^2.
\end{equation}

Latent consistency compares the rolled-out latent states with states obtained
by re-encoding shifted windows that include the observed future. In the
disentangled model, this consistency term is applied to the degradation
component:
\begin{equation}
  \mathcal{L}_{\mathrm{latent}}
  =
  \frac{1}{H}\sum_{h=1}^{H}
  \left\|
    z_{t+h}^{\mathrm{deg}}
    -
    \bar{z}_{t+h}^{\mathrm{deg}}
  \right\|_2^2,
\end{equation}
where \(\bar{z}_{t+h}^{\mathrm{deg}}\) is produced by encoding the shifted
window ending at \(t+h\), and gradients are stopped through this target. The
monotonicity loss discourages decreasing degradation risk during the rollout:
\begin{equation}
  \mathcal{L}_{\mathrm{mono}}
  =
  \frac{1}{H}\sum_{h=1}^{H}
  \max(0, -(r_{t+h} - r_{t+h-1})).
\end{equation}

To make the condition component carry operating context, an auxiliary head
predicts the future setting:
\begin{equation}
  \mathcal{L}_{\mathrm{cond}}
  =
  \frac{1}{H}\sum_{h=1}^{H}
  \left\|
    \hat{a}_{t+h} - a_{t+h}
  \right\|_2^2.
\end{equation}
The decorrelation term penalizes empirical covariance between degradation and
condition states, and between degradation states and operating settings:
\begin{equation}
  \mathcal{L}_{\mathrm{decor}}
  =
  \left\|
    \operatorname{Cov}(z^{\mathrm{deg}}, z^{\mathrm{cond}})
  \right\|_F^2
  +
  \left\|
    \operatorname{Cov}(z^{\mathrm{deg}}, a)
  \right\|_F^2.
\end{equation}
Finally, the smoothness penalty controls the average magnitude of the liquid
state update:
\begin{equation}
  \mathcal{L}_{\mathrm{smooth}}
  =
  \frac{1}{H}\sum_{h=1}^{H}
  \left\|
    \Delta z_{t+h}^{\mathrm{deg}}
  \right\|_2.
\end{equation}

\subsection{Implementation Details}

The encoder is a GRU over the concatenated operating settings and sensor
measurements. The liquid transition, decoder, RUL head, and condition head are
implemented as small multilayer perceptrons with smooth nonlinearities. The
disentangled model splits the latent dimension approximately in half between
degradation and condition components. Training uses teacher-observed future
windows only to form latent-consistency targets; the rollout itself predicts
future sensors autoregressively through the learned latent transition. Gradients
through the latent-consistency targets are stopped, so this term regularizes the
rollout toward the encoder geometry without collapsing the encoder to the
transition output.

\begin{algorithm}[t]
\caption{Training Disentangled Liquid Latent Dynamics}
\label{alg:liquid-latent}
\begin{algorithmic}[1]
\Require History window \(X_t=[u_{t-L+1},\ldots,u_t]\), future settings
\(a_{t+1:t+H}\), future sensors \(x_{t+1:t+H}\), and target
\(\mathrm{RUL}_{t+H}\)
\Ensure Trained encoder \(E_\theta\), liquid transition \(G_\psi\), condition
update \(C_\omega\), decoder \(D_\phi\), setting head \(P_\rho\), and RUL head
\(Q_\eta\)
\State Encode history:
\((z_t^{\mathrm{deg}}, z_t^{\mathrm{cond}}) \gets E_\theta(X_t)\)
\State Initialize risk \(r_t \gets 1 - Q_\eta(z_t^{\mathrm{deg}})\)
\For{\(h=1,\ldots,H\)}
  \State \(z_{t+h}^{\mathrm{cond}} \gets
  C_\omega(z_{t+h-1}^{\mathrm{cond}}, a_{t+h})\)
  \State \((z_{t+h}^{\mathrm{deg}}, \Delta z_{t+h}^{\mathrm{deg}})
  \gets G_\psi(z_{t+h-1}^{\mathrm{deg}},
  [a_{t+h}, z_{t+h}^{\mathrm{cond}}])\)
  \State \(\hat{x}_{t+h} \gets
  D_\phi(z_{t+h}^{\mathrm{deg}}, z_{t+h}^{\mathrm{cond}})\)
  \State \(r_{t+h} \gets 1 - Q_\eta(z_{t+h}^{\mathrm{deg}})\)
  \State \(\hat{a}_{t+h} \gets P_\rho(z_{t+h}^{\mathrm{cond}})\)
  \State Re-encode the shifted observed window ending at \(t+h\) to obtain
  stop-gradient target \(\bar{z}_{t+h}^{\mathrm{deg}}\)
\EndFor
\State Compute the weighted objective in Eq.~\ref{eq:total-loss}
\State Update all trainable parameters by backpropagation
\end{algorithmic}
\end{algorithm}

\section{Experiments}

\subsection{Dataset}

We evaluate on the NASA C-MAPSS turbofan degradation benchmark
\citep{saxena2008damage}. The benchmark contains simulated run-to-failure
trajectories of aircraft engines. Each cycle contains three operating settings
and 21 sensor measurements. The four subsets differ in the number of operating
conditions and fault modes: FD001 contains one operating condition and one fault
mode; FD002 contains multiple operating conditions and one fault mode; FD003
contains one operating condition and multiple fault modes; and FD004 contains
multiple operating conditions and multiple fault modes. This structure is
important for our study because FD002 and FD004 explicitly test whether the
model can handle operating-condition variation while learning degradation
dynamics.

For each engine trajectory, the raw RUL at cycle \(t\) is computed as the
number of remaining cycles before the final observed cycle of that engine. We
use the common capped RUL target,
\begin{equation}
  \mathrm{RUL}^{\mathrm{cap}}_t
  =
  \frac{\min(\mathrm{RUL}_t, 125)}{125},
\end{equation}
so the supervised RUL target lies in \([0,1]\). Sensor and operating-setting
features are standardized using statistics computed only from the training
engines of the corresponding subset.

We construct supervised windows from each trajectory. A history window of
length \(L=30\) is used as input, and the model predicts the next \(H=5\)
cycles. The input at each history step is the concatenation of operating
settings and sensor values. The future operating settings
\(a_{t+1:t+H}\) are provided to the model during rollout, and the prediction
target is the future sensor sequence \(x_{t+1:t+H}\) together with the capped
RUL at the end of the horizon. For each subset and random seed, engines are
split by unit identity into 70\% training, 15\% validation, and 15\% test
units. This unit-level split prevents windows from the same engine trajectory
from appearing in both training and evaluation.

\subsection{Evaluation Protocol}

We evaluate three complementary aspects of the model: sensor forecasting,
direct RUL regression, and latent degradation alignment. Sensor forecasting is
measured by root mean squared error (RMSE) over all predicted sensors and all
forecasting horizons:
\begin{equation}
  \mathrm{RMSE}_{\mathrm{sensor}}
  =
  \sqrt{
    \frac{1}{N H d_x}
    \sum_{i=1}^{N}
    \sum_{h=1}^{H}
    \left\|
      \hat{x}^{(i)}_{t_i+h} - x^{(i)}_{t_i+h}
    \right\|_2^2
  }.
\end{equation}
We also report first-step and last-step sensor RMSE when analyzing horizon
effects. RUL regression is measured by RMSE after rescaling the normalized
prediction back to cycle units:
\begin{equation}
  \mathrm{RMSE}_{\mathrm{RUL}}
  =
  \sqrt{
    \frac{1}{N}
    \sum_{i=1}^{N}
    \left(
      \widehat{\mathrm{RUL}}^{(i)} -
      \mathrm{RUL}^{(i)}
    \right)^2
  }.
\end{equation}

To evaluate whether the learned state reflects degradation progress, we compute
rank correlations against a capped degradation variable,
\begin{equation}
  d_t = 125 - \min(\mathrm{RUL}_t, 125).
\end{equation}
First, we measure the Spearman correlation between predicted risk
\(r_t = 1-\widehat{\mathrm{RUL}}_t\) and \(d_t\). Second, for liquid models, we
measure the Spearman correlation between the average rollout speed
\(\frac{1}{H}\sum_{h=1}^{H}\|\Delta z_{t+h}\|_2\) and \(d_t\). The latter
metric is not a conventional RUL accuracy metric; it is intended to test
whether the learned latent dynamics changes more strongly as the engine moves
toward failure.

As an auxiliary degradation-detection evaluation, we convert RUL into a binary
normal/abnormal label using two separated regions: samples with
\(\mathrm{RUL}\geq 125\) are treated as normal, and samples with
\(\mathrm{RUL}\leq 30\) are treated as abnormal. Intermediate samples are
excluded from this binary evaluation. The detection score is the model risk
\(r_t\). We report AUROC, AUPRC, balanced accuracy, and F1. The threshold for
balanced accuracy and F1 is selected on the validation split by maximizing F1
and is then applied once to the held-out test split.

All reported experiments use five random seeds,
\(\{7, 11, 13, 17, 23\}\), and train for 50 epochs. The validation criterion is
the sum of sensor RMSE and a small weighted RUL RMSE term,
\(\mathrm{RMSE}_{\mathrm{sensor}} + 0.02\,\mathrm{RMSE}_{\mathrm{RUL}}\), which
selects checkpoints that preserve forecasting accuracy while discouraging
pathological RUL predictions. The primary comparison is against a GRU baseline
with the same history window and future-setting inputs. We also compare basic
liquid latent dynamics, disentangled liquid dynamics, and the full
disentangled model with RUL, monotonicity, latent-consistency, condition, and
decorrelation losses.

\subsection{Comparison with Existing Baselines}

Table~\ref{tab:sensor-rul-comparison} compares the current GRU baseline and the
full disentangled liquid model on sensor forecasting and RUL regression. The
results reveal a useful but nontrivial trade-off. The proposed model gives clear
sensor-forecasting gains on the multi-condition subsets FD002 and FD004, where
operating-condition variation is most prominent. On FD002, sensor RMSE decreases
from 0.1058 to 0.0627; on FD004, it decreases from 0.0936 to 0.0625. These are
the settings where disentangling degradation from condition variation should be
most beneficial. However, the GRU remains stronger for direct RUL RMSE on three
of the four subsets. This supports our central interpretation: the proposed
model is better positioned as a latent degradation dynamics model and sensor
world model than as a purely optimized RUL regressor.

\begin{table}[t]
  \centering
  \small
  \caption{Sensor forecasting and RUL regression comparison on C-MAPSS. Lower
  is better for sensor RMSE and RUL RMSE; higher is better for Spearman
  correlations.}
  \label{tab:sensor-rul-comparison}
  \begin{tabular}{lrrrrr}
    \toprule
    Subset & GRU sensor & Dis+RUL sensor & GRU RUL & Dis+RUL RUL & Speed \(\rho\) \\
    \midrule
    FD001 & 0.4401 & 0.4415 & 15.4393 & 16.2309 & 0.5533 \\
    FD002 & 0.1058 & \textbf{0.0627} & \textbf{18.6980} & 19.5579 & 0.6115 \\
    FD003 & \textbf{0.3357} & 0.3398 & \textbf{14.3790} & 14.6824 & 0.5859 \\
    FD004 & 0.0936 & \textbf{0.0625} & \textbf{19.7048} & 21.1477 & 0.6335 \\
    \bottomrule
  \end{tabular}
\end{table}

\begin{figure}[t]
  \centering
  \includegraphics[width=\linewidth]{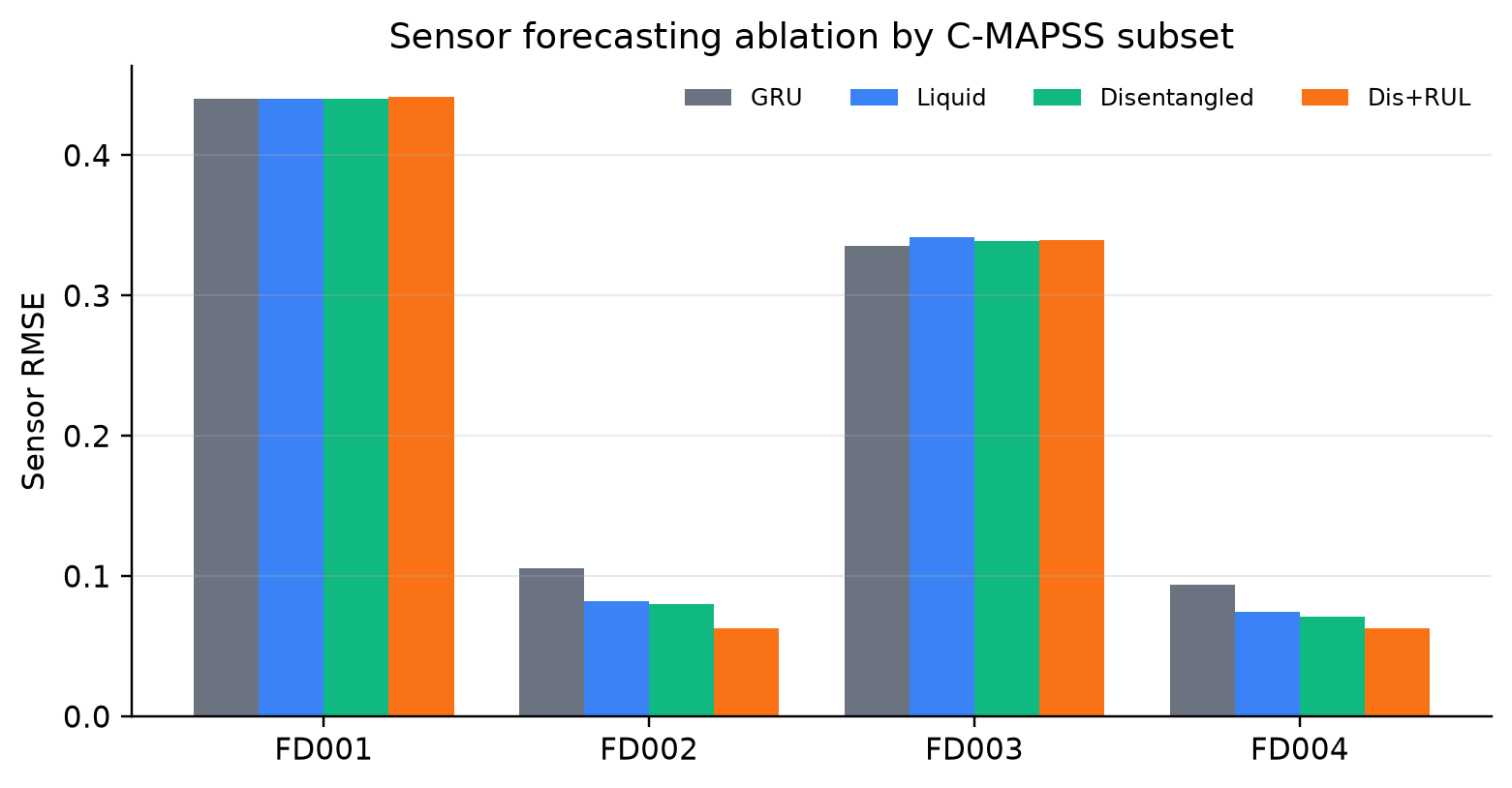}
  \caption{Sensor forecasting RMSE by subset. The proposed Dis+RUL model
  provides the clearest improvement on the multi-condition subsets FD002 and
  FD004.}
  \label{fig:sensor-rmse-ablation}
\end{figure}

Figure~\ref{fig:sensor-rmse-ablation} visualizes the same forecasting pattern.
The gains are small or absent on FD001 and FD003, but become large on FD002 and
FD004. This is consistent with the motivation of the method: a separate
condition-modulated latent dynamics is most useful when operating regimes vary.

We further compare against a broader set of degradation-detection baselines in
Table~\ref{tab:detection-comparison}. This comparison uses the normal/abnormal
protocol described above, with risk as the detection score. Under this protocol,
both the current GRU and the proposed Dis+RUL model are very strong, and the
task is close to saturated. The proposed model obtains the best AUPRC, balanced
accuracy, and F1 among the compared methods, while the GRU has a marginally
higher risk-degradation rank correlation. Compared with the strongest previous
non-current baseline, \texttt{knn\_last\_5}, Dis+RUL improves AUPRC from 0.9921
to 0.9984, balanced accuracy from 0.9717 to 0.9953, and F1 from 0.9592 to
0.9923. The practical implication is not that detection alone proves a superior
RUL model; rather, it shows that the learned risk signal is highly separable for
early-vs-late degradation while the latent dynamics model simultaneously
improves sensor forecasting.

\begin{table}[t]
  \centering
  \small
  \caption{Degradation detection comparison with existing baselines. Higher is
  better for all metrics. Current models use \(r_t=1-\widehat{\mathrm{RUL}}_t\)
  as the detection score.}
  \label{tab:detection-comparison}
  \begin{tabular}{lrrrrr}
    \toprule
    Method & AUROC & AUPRC & BAcc & F1 & \(\rho\) \\
    \midrule
    GRU-current & \textbf{0.9997} & 0.9978 & 0.9948 & 0.9921 & \textbf{0.8570} \\
    Dis+RUL-current & 0.9997 & \textbf{0.9984} & \textbf{0.9953} & \textbf{0.9923} & 0.8564 \\
    \texttt{knn\_last\_5} & 0.9951 & 0.9921 & 0.9717 & 0.9592 & 0.6831 \\
    Normal-World Ours & 0.9866 & 0.9712 & 0.9594 & 0.9316 & 0.7056 \\
    \texttt{gdn\_lite} & 0.9184 & 0.8969 & 0.8704 & 0.8089 & 0.5159 \\
    \texttt{pca\_last\_8} & 0.8322 & 0.7587 & 0.7623 & 0.6526 & 0.3848 \\
    \texttt{gru\_forecast} & 0.7632 & 0.6500 & 0.7463 & 0.6914 & 0.3143 \\
    \texttt{tcn\_forecast} & 0.7582 & 0.6425 & 0.7397 & 0.6799 & 0.2872 \\
    \texttt{gaussian\_last} & 0.7527 & 0.6924 & 0.7728 & 0.7159 & 0.3906 \\
    \texttt{transformer\_forecast} & 0.7525 & 0.6496 & 0.7516 & 0.7063 & 0.3150 \\
    \texttt{ridge\_predictor} & 0.7086 & 0.5636 & 0.6781 & 0.5570 & 0.2282 \\
    \bottomrule
  \end{tabular}
\end{table}

\begin{figure}[t]
  \centering
  \includegraphics[width=0.92\linewidth]{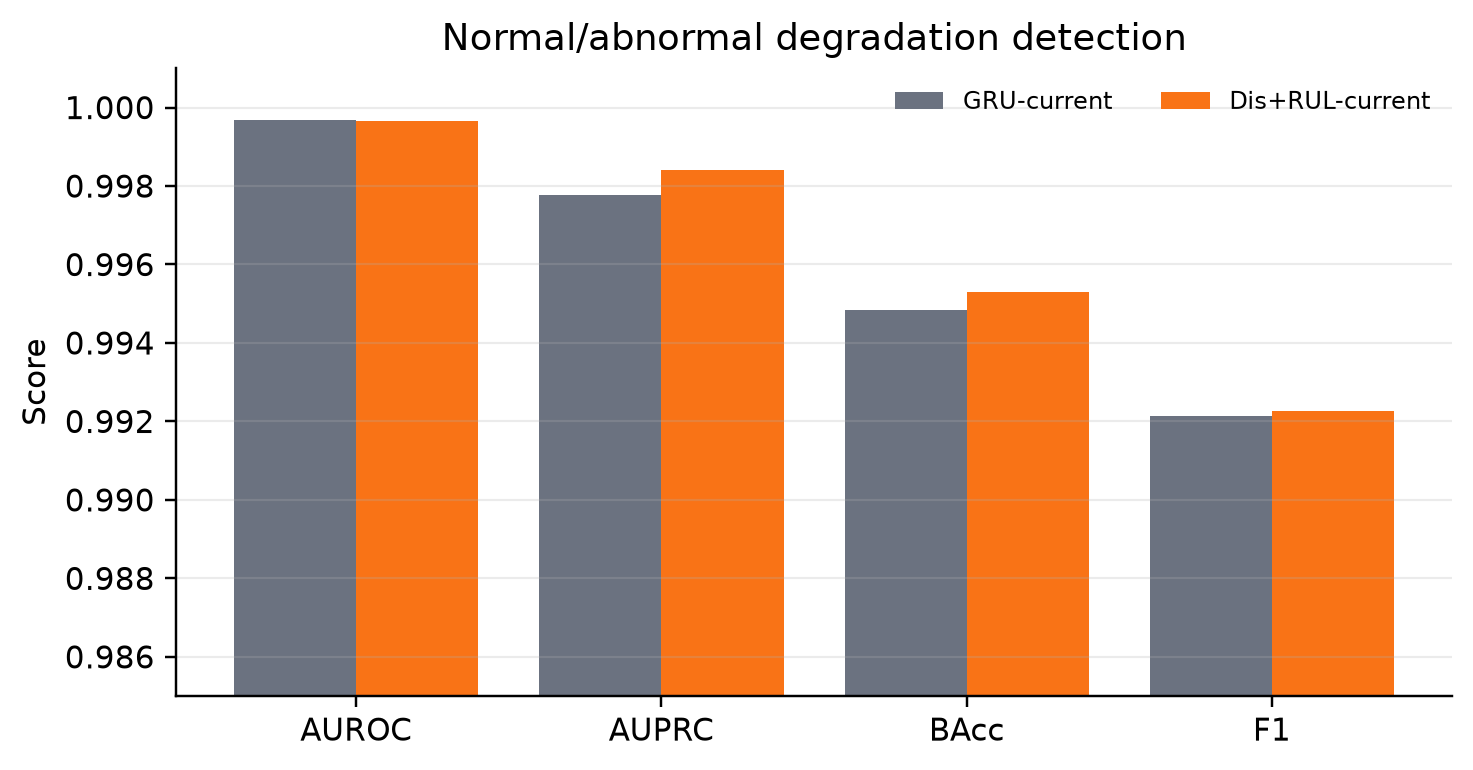}
  \caption{Normal/abnormal degradation detection metrics for current models.
  Detection is close to saturated under the selected split; Dis+RUL gives small
  but consistent gains in AUPRC, balanced accuracy, and F1.}
  \label{fig:detection-bars}
\end{figure}

Figure~\ref{fig:detection-bars} highlights the near-saturated detection
performance of both current models. Dis+RUL is slightly better on AUPRC,
balanced accuracy, and F1, while AUROC is essentially tied. This supports the
view that the learned risk score is highly effective for separating normal and
late-degradation samples, but it also warns against overinterpreting detection
as full RUL calibration.

The detection results should be interpreted with care. The normal/abnormal
split removes intermediate degradation stages, so it is easier than calibrated
RUL regression over the full trajectory. For this reason, we use the detection
comparison as evidence that the learned risk score is discriminative, while the
RUL RMSE results remain the stricter test of lifetime calibration. Under this
more conservative reading, the main empirical claim is that Dis+RUL combines
strong degradation detection with better sensor forecasting and an inspectable
latent degradation trajectory, but does not yet dominate direct GRU regression
on RUL RMSE.

\subsection{Ablation Study}

Table~\ref{tab:ablation-overall} summarizes the main ablation over model
structure and supervision. Moving from the GRU baseline to basic liquid latent
dynamics improves overall sensor RMSE from 0.2438 to 0.2347, suggesting that
rolling a latent state forward is useful for multistep sensor forecasting.
Adding the degradation/condition factorization further improves sensor RMSE to
0.2325. The full Dis+RUL model, which adds RUL supervision, monotonic risk,
latent consistency, condition prediction, and decorrelation, gives the best
overall sensor RMSE of 0.2266. The most important change is in the latent-speed
alignment: the average speed-vs-degradation Spearman correlation increases from
0.2847 for basic liquid dynamics and 0.2966 for disentangled liquid dynamics to
0.5960 for Dis+RUL.

This ablation clarifies the role of each design choice. The liquid transition
alone improves forecasting because it gives the model an explicit rollout
mechanism, but its latent increments are not yet strongly ordered by
degradation. The disentangled state improves forecasting slightly by separating
condition-dependent variation from the state used for degradation prediction,
but this structural split alone is also insufficient to make the trajectory
monotonic or health-aligned. The large increase in speed \(\rho\) appears only
after adding degradation-specific supervision and latent consistency. This
means that the state-change signal \(\Delta z^{\mathrm{deg}}\) is learned
through the combination of rollout prediction, re-encoding consistency, and
monotonic risk, rather than emerging automatically from the liquid architecture
alone.

\begin{table}[t]
  \centering
  \small
  \caption{Overall ablation across all four C-MAPSS subsets. Lower is better
  for sensor and RUL RMSE; higher is better for rank correlations. Speed
  \(\rho\) measures whether the magnitude of latent-state change aligns with
  degradation progress.}
  \label{tab:ablation-overall}
  \begin{tabular}{lrrrr}
    \toprule
    Model & Sensor RMSE & RUL RMSE & Risk \(\rho\) & Speed \(\rho\) \\
    \midrule
    GRU & 0.2438 & \textbf{17.0553} & \textbf{0.8636} & -- \\
    Liquid & 0.2347 & 17.2997 & 0.8586 & 0.2847 \\
    Disentangled & 0.2325 & 17.4737 & 0.8531 & 0.2966 \\
    Dis+RUL & \textbf{0.2266} & 17.9047 & 0.8520 & \textbf{0.5960} \\
    \bottomrule
  \end{tabular}
\end{table}

The subset-level ablation in Table~\ref{tab:ablation-subset} shows that the
forecasting benefit is concentrated on the multi-condition subsets. On FD002,
Dis+RUL reduces sensor RMSE by 40.7\% relative to GRU, and on FD004 it reduces
sensor RMSE by 33.2\%. These are precisely the subsets where operating
conditions can obscure degradation. On FD001 and FD003, where the operating
condition structure is simpler, the liquid models do not improve forecasting
over the GRU baseline. This pattern gives a scoped interpretation of the
method: it is not uniformly better on every subset, but is most useful in the
setting it targets, namely degradation modeling under condition variation.

The same pattern appears in the state-evolution metric. In FD002 and FD004,
basic liquid dynamics has near-zero speed correlation, indicating that the
latent state may move, but its movement is not aligned with health progression.
After adding the full degradation-specific objective, speed \(\rho\) reaches
0.6115 on FD002 and 0.6335 on FD004. Thus, the ablation supports a two-part
claim: liquid rollout helps prediction, while the RUL, monotonic, and
latent-consistency losses make the rollout direction meaningful as a
degradation-state transition.

\begin{table}[t]
  \centering
  \small
  \caption{Sensor forecasting ablation by subset. Lower is better. The largest
  gains appear on FD002 and FD004, the multi-condition subsets.}
  \label{tab:ablation-subset}
  \begin{tabular}{lrrrr}
    \toprule
    Subset & GRU & Liquid & Disentangled & Dis+RUL \\
    \midrule
    FD001 & \textbf{0.4401} & 0.4402 & 0.4402 & 0.4415 \\
    FD002 & 0.1058 & 0.0822 & 0.0799 & \textbf{0.0627} \\
    FD003 & \textbf{0.3357} & 0.3419 & 0.3390 & 0.3398 \\
    FD004 & 0.0936 & 0.0743 & 0.0711 & \textbf{0.0625} \\
    Overall & 0.2438 & 0.2347 & 0.2325 & \textbf{0.2266} \\
    \bottomrule
  \end{tabular}
\end{table}

\begin{figure}[t]
  \centering
  \includegraphics[width=\linewidth]{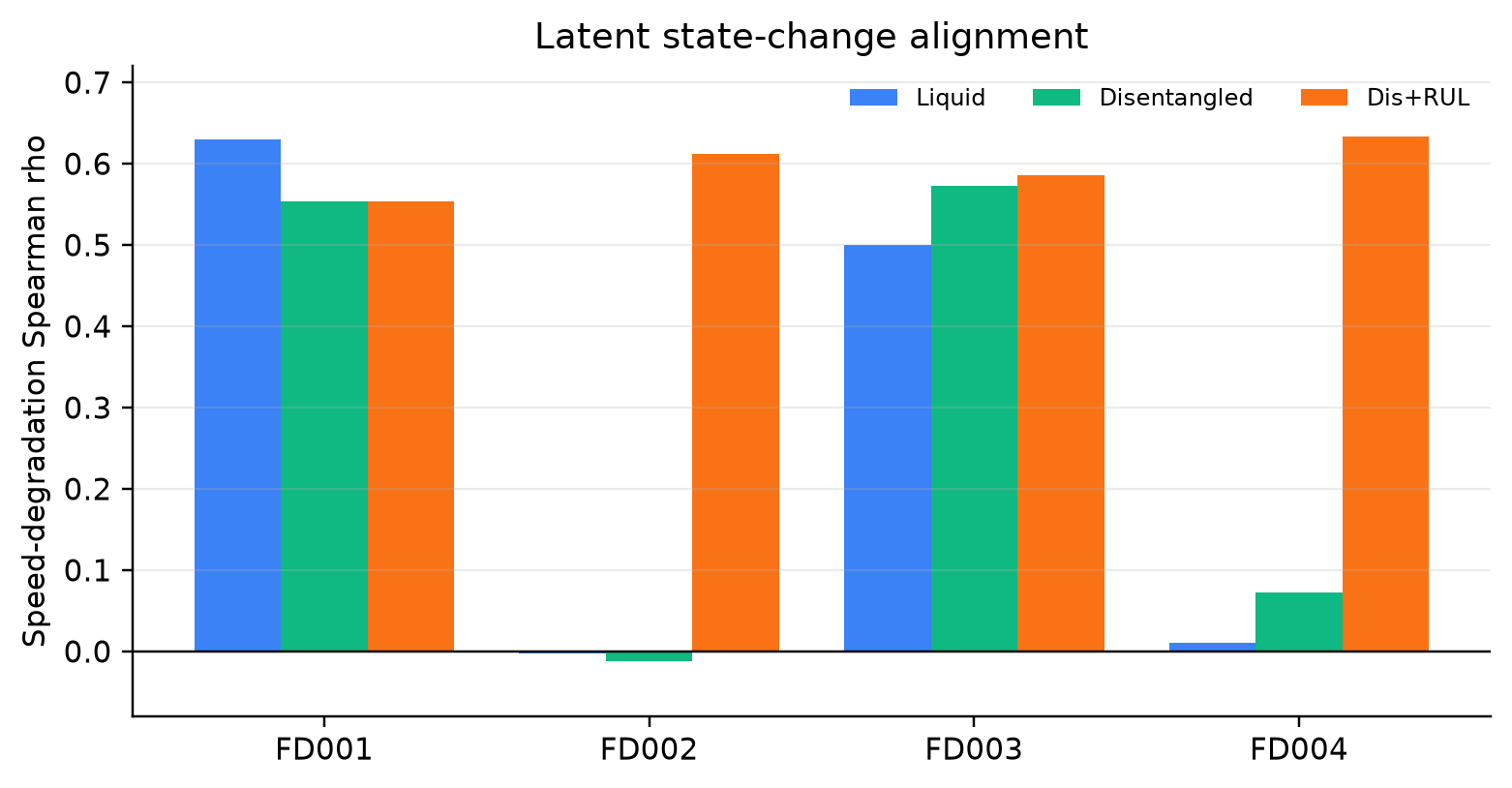}
  \caption{Latent state-change alignment by subset. Speed \(\rho\) measures the
  Spearman correlation between degradation and the average magnitude of
  \(\Delta z\). The full Dis+RUL objective strongly increases this alignment on
  FD002 and FD004.}
  \label{fig:speed-alignment}
\end{figure}

Figure~\ref{fig:speed-alignment} makes the ablation effect on state evolution
more explicit. On FD002 and FD004, the basic liquid model has almost no
positive speed-degradation alignment, which means that merely rolling a latent
state forward is not enough. The full Dis+RUL objective changes this behavior:
the learned update magnitude becomes strongly associated with degradation
progress. This is one of the clearest quantitative signs that the model learns a
meaningful state-change trend.

\subsection{Analysis of Learned State Evolution}

The ablation results indicate that the full objective does more than improve
sensor forecasting. It changes the geometry and dynamics of the learned latent
state. In the basic liquid and disentangled variants, the latent update speed is
only weakly aligned with degradation progress, especially on FD002 and FD004.
For example, the speed correlation on FD004 is 0.0111 for the basic liquid
model and 0.0720 for the disentangled model, but rises to 0.6335 after adding
RUL, monotonicity, and degradation-state latent consistency. Similarly, on FD002
the speed correlation changes from approximately zero to 0.6115. This suggests
that the added supervision does not merely alter the final RUL head; it makes
the latent transition itself move in a way that is ordered by degradation.

\begin{figure}[t]
  \centering
  \includegraphics[width=\linewidth]{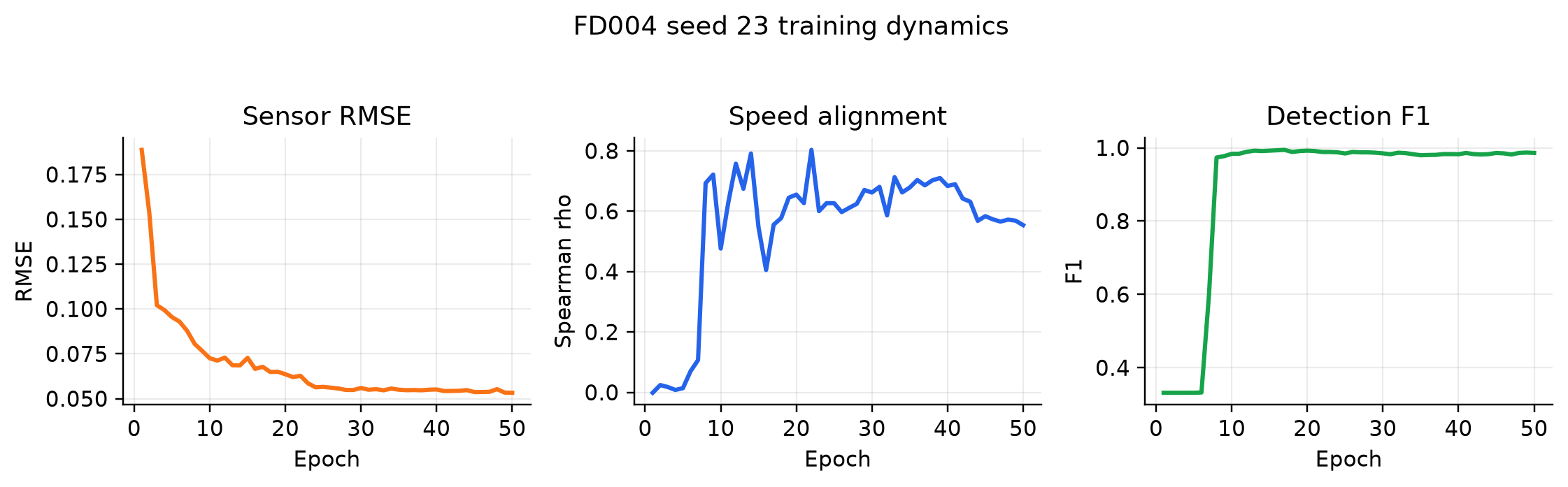}
  \caption{Training dynamics for FD004 seed 23. Forecasting error decreases,
  detection F1 becomes high, and speed-degradation alignment rises from near
  zero to a strong positive value, indicating that the learned state increments
  become ordered by degradation during training.}
  \label{fig:training-dynamics}
\end{figure}

Training dynamics on FD004 provide another view of this process
(Fig.~\ref{fig:training-dynamics}). Sensor RMSE decreases rapidly during early
epochs, while detection F1 also becomes high after the risk head learns a
separable normal/abnormal score. The speed-alignment curve is more informative
for our main claim: it rises from near zero to a high positive correlation,
showing that the latent transition gradually becomes organized with respect to
degradation. Thus, the model first learns to forecast and separate risk, and
then stabilizes a latent trajectory whose increments carry degradation
information.

\begin{figure}[t]
  \centering
  \begin{minipage}{0.48\linewidth}
    \centering
    \includegraphics[width=\linewidth]{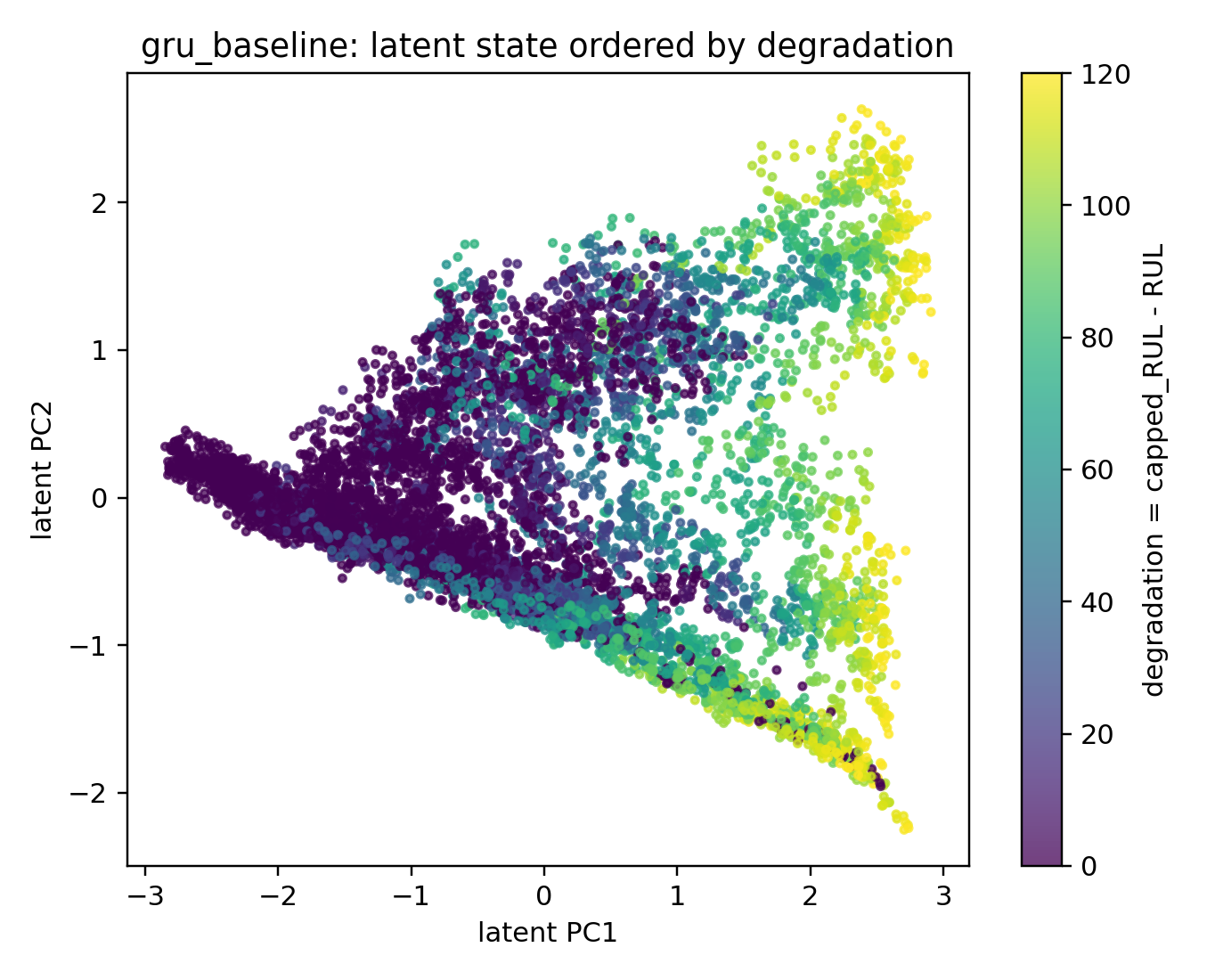}
    \centerline{\small (a) GRU hidden state}
  \end{minipage}
  \hfill
  \begin{minipage}{0.48\linewidth}
    \centering
    \includegraphics[width=\linewidth]{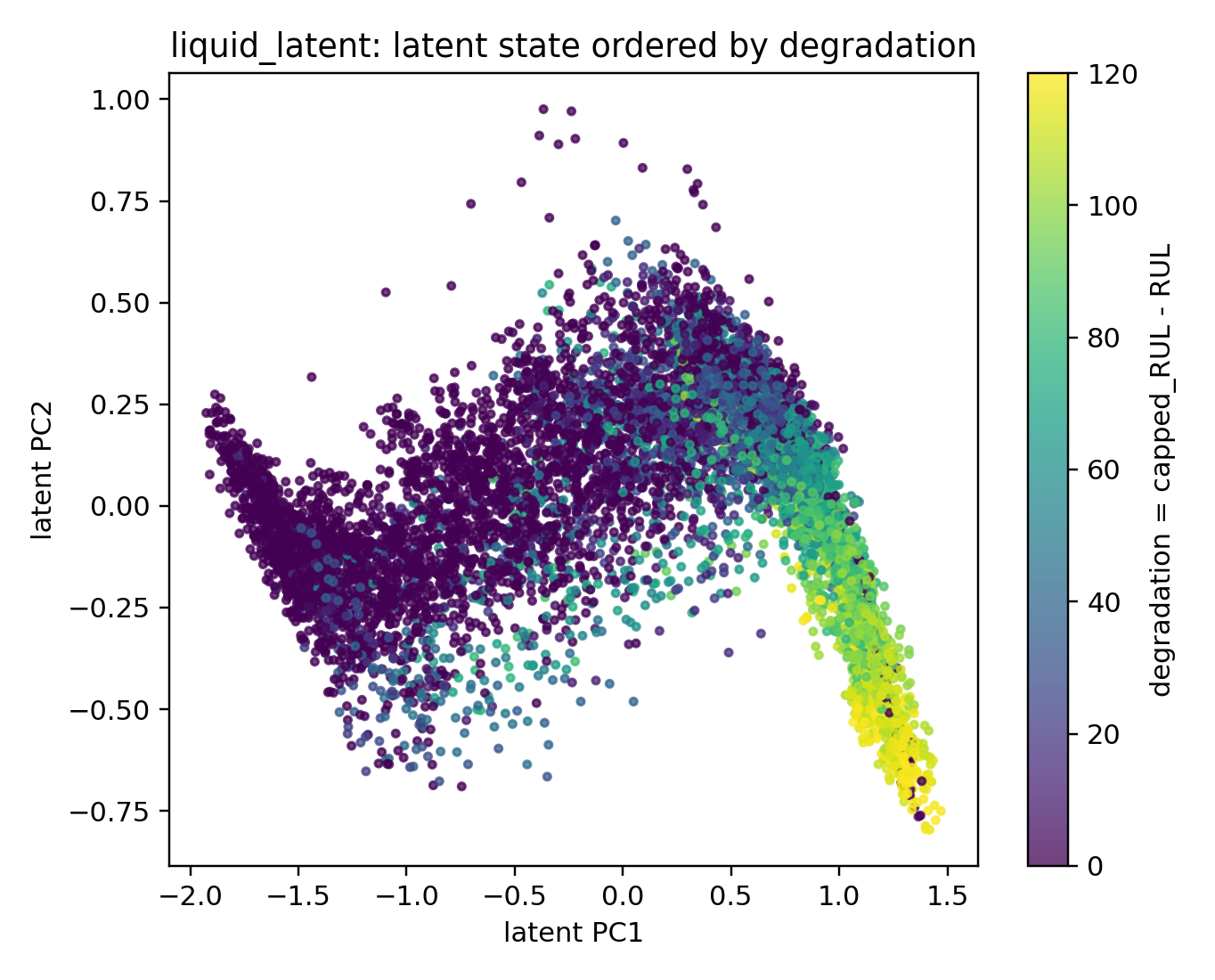}
    \centerline{\small (b) Full liquid latent state}
  \end{minipage}
  \vspace{0.7em}

  \begin{minipage}{0.48\linewidth}
    \centering
    \includegraphics[width=\linewidth]{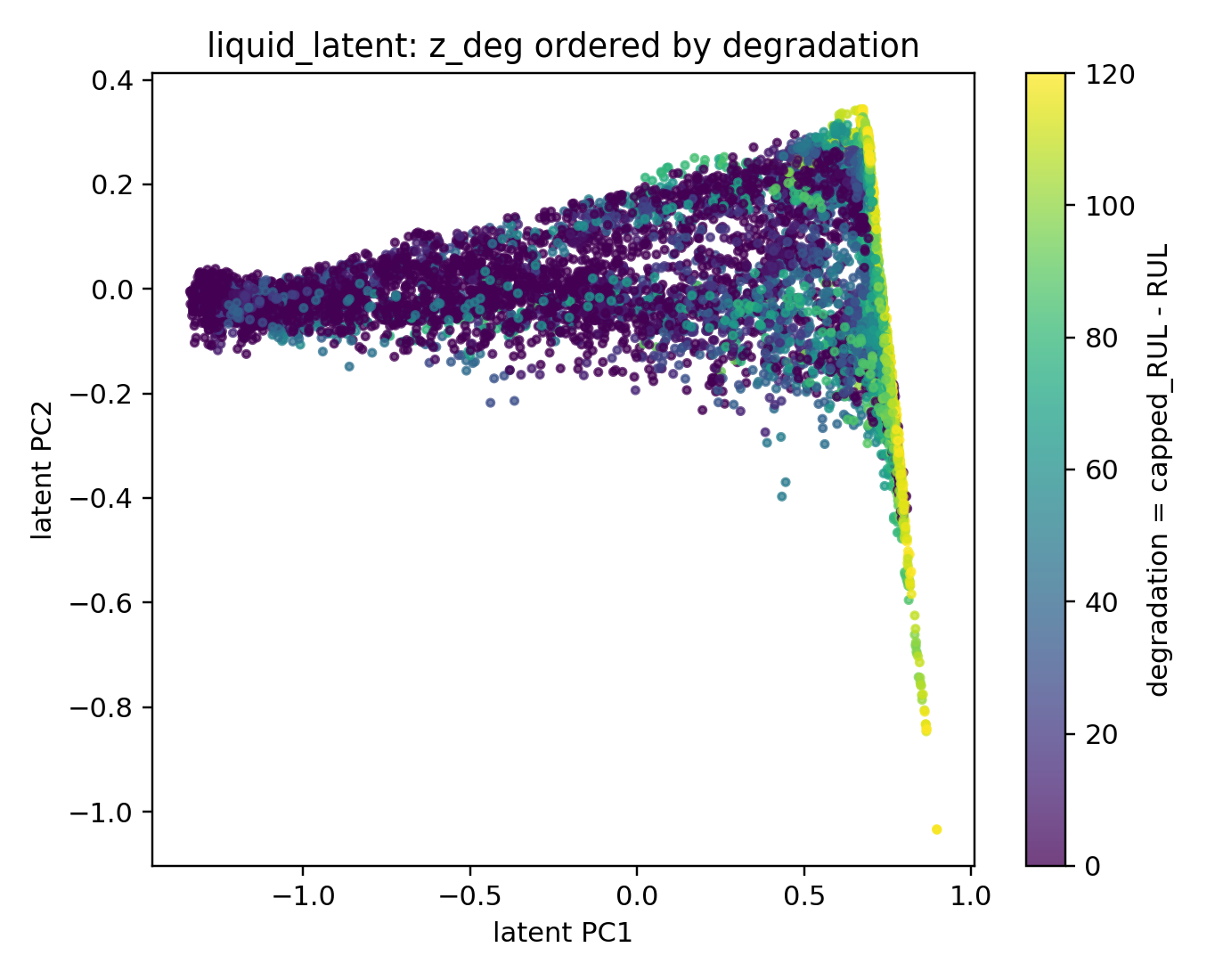}
    \centerline{\small (c) Degradation state \(z^{\mathrm{deg}}\)}
  \end{minipage}
  \hfill
  \begin{minipage}{0.48\linewidth}
    \centering
    \includegraphics[width=\linewidth]{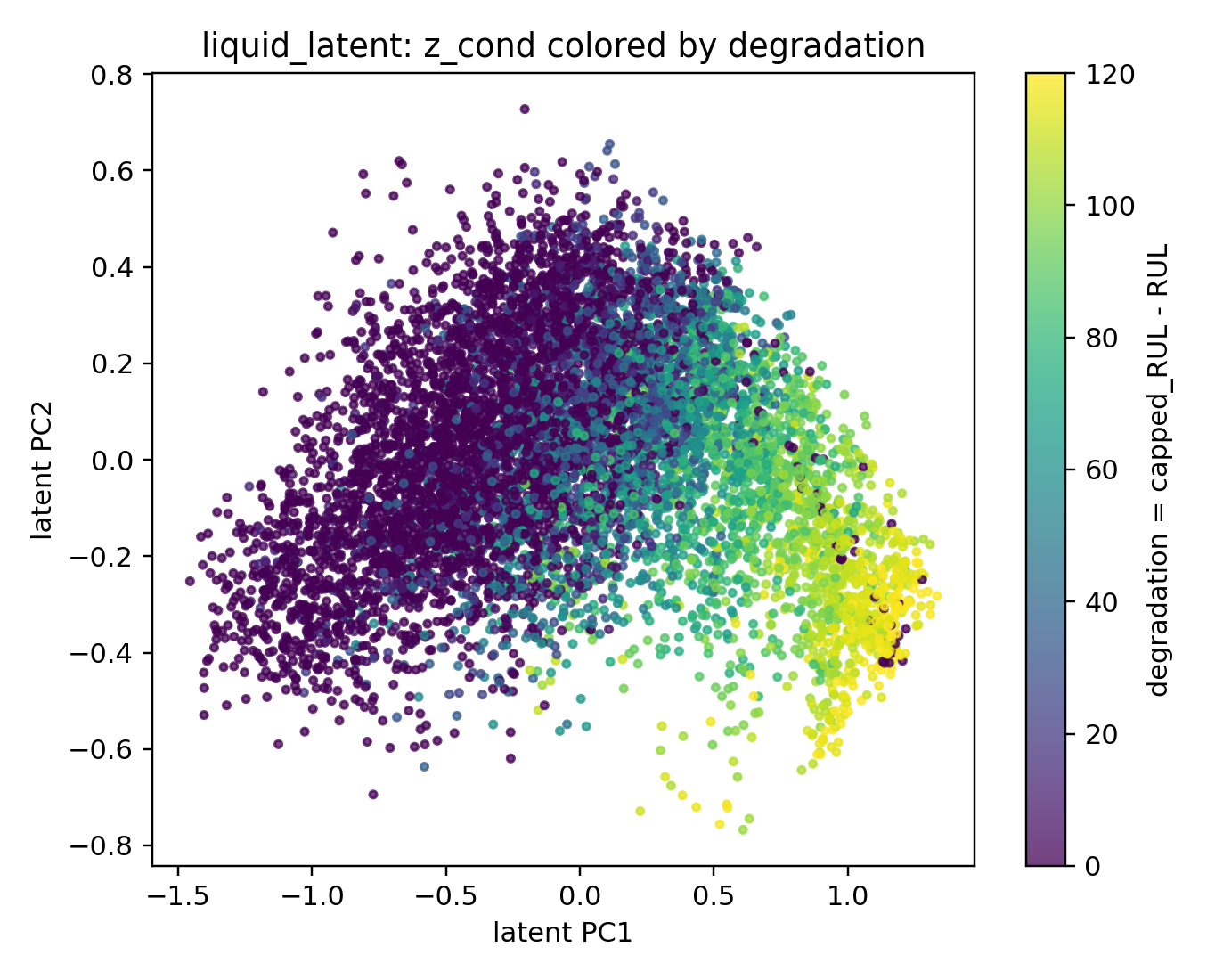}
    \centerline{\small (d) Condition state \(z^{\mathrm{cond}}\)}
  \end{minipage}
  \caption{PCA visualization of learned states on FD004 with seed 23, colored
  by capped degradation. Compared with the GRU hidden state, the proposed model
  learns a more ordered latent evolution. The degradation component
  \(z^{\mathrm{deg}}\) shows the clearest progression from early states to
  terminal degraded states, while \(z^{\mathrm{cond}}\) remains more diffuse.}
  \label{fig:latent-pca}
\end{figure}

Figure~\ref{fig:latent-pca} provides a qualitative view of this effect on FD004
with seed 23. The GRU baseline latent PCA does not expose a clean degradation
coordinate: early, middle, and late states are mixed in a representation trained
primarily for direct prediction. The full liquid latent state is more ordered,
but it still contains both degradation and condition information. When the
state is decomposed, \(z^{\mathrm{deg}}\) forms the clearest degradation
trajectory: early states occupy a broad region on one side of the projection,
while high-degradation states concentrate along a terminal band. In contrast,
\(z^{\mathrm{cond}}\) is more diffuse and does not form the same clean terminal
degradation structure. This difference supports the intended factorization, but
also reveals its limitation: \(z^{\mathrm{cond}}\) is not completely free of
degradation information, so the disentanglement should be viewed as useful but
imperfect.

Panel (c) in Fig.~\ref{fig:latent-pca} provides the most direct visual evidence.
It visualizes only
\(z^{\mathrm{deg}}\), the subspace that receives RUL, monotonicity, and
latent-consistency supervision. If the model were merely using
\(z^{\mathrm{deg}}\) as another unconstrained feature vector, the projection
would look similar to the GRU hidden state or to the diffuse condition state.
Instead, the samples organize along a visible degradation direction: low
degradation states are spread over the early region, intermediate states move
through the middle of the projection, and high degradation states concentrate
near a terminal band. This geometry matches the intended behavior of the
degradation state. It suggests that \(z^{\mathrm{deg}}\) is not only
predictive, but also behaves like a state coordinate whose position changes as
the engine moves toward failure.

This visual trend is consistent with the learned liquid increment. During
rollout, the model repeatedly applies
\(z_{t+h}^{\mathrm{deg}}=z_{t+h-1}^{\mathrm{deg}}+\Delta
z_{t+h}^{\mathrm{deg}}\). The ordered structure in Fig.~\ref{fig:latent-pca}(c)
therefore indicates that the accumulated increments move the latent state along
a degradation-oriented path rather than scattering it arbitrarily. The high
speed-vs-degradation correlations provide the quantitative counterpart: as
degradation increases, the magnitude of the learned transition
\(\|\Delta z^{\mathrm{deg}}\|\) becomes more aligned with the degradation
level. Together, panel (c) and the speed correlations form the main evidence
that the model learns a trend-changing hidden state.

Together, the quantitative speed correlations and the PCA visualizations
support the claim that the model learns a degradation-related state evolution
rather than only a static predictive embedding. The evidence is strongest on
FD002 and FD004, where operating-condition variation makes a single hidden state
harder to interpret. At the same time, the weaker RUL RMSE and residual
degradation leakage into \(z^{\mathrm{cond}}\) show that the learned state is
not yet a perfectly calibrated health coordinate. This is the main remaining
gap between an interpretable latent dynamics model and a fully accurate
prognostic estimator.

\section{Future Work}

This work suggests several directions for improvement. First, the current model
learns a degradation-oriented latent trajectory, but its RUL calibration is
still weaker than the GRU baseline. Future work should separate representation
learning from calibration more explicitly, for example by adding calibration
losses, uncertainty estimates, or post-hoc monotone calibration on top of the
learned degradation coordinate. Second, the disentanglement between
\(z^{\mathrm{deg}}\) and \(z^{\mathrm{cond}}\) is useful but imperfect. Stronger
decorrelation, adversarial condition removal, or condition-invariant contrastive
objectives may reduce degradation leakage into the condition state. Third, the
current evaluation uses C-MAPSS, where degradation trajectories are simulated
and relatively clean. Real industrial systems may contain maintenance events,
sensor drift, missing observations, and non-monotone operating regimes, all of
which require more robust latent dynamics.

A broader direction is to extend liquid latent state dynamics from machinery
degradation to biological and medical progression. Many diseases can also be
viewed as partially observed dynamical processes: clinical measurements,
laboratory tests, imaging biomarkers, or wearable signals are observations of an
underlying health state that changes over time. Related work on time-domain
event recognition, clinical temporal modeling, medical prior knowledge, and
causal diagnosis suggests that progression modeling benefits from explicitly
representing temporal change and domain constraints
\citep{su2019pooled,nie2018mitosis,nie2023icu,li2023brain,nie2024t2td,li2024cisepsis,zhang2025cabg}.
A model that learns \(\Delta z\), rather than only a static risk score, could
potentially describe how a patient's latent disease state evolves and how
interventions or conditions modulate that evolution. This extension should be
treated carefully. Unlike C-MAPSS, disease progression involves heterogeneous
populations, irregular sampling, treatment effects, confounding, and
high-stakes clinical decisions. Applying this framework to medicine will
require uncertainty-aware prediction, causal validation, clinically meaningful
state definitions, and prospective evaluation. Nevertheless, the present
results motivate studying liquid latent dynamics as a tool for
interpretable progression modeling in domains where the main object of interest
is an unobserved state trajectory.

\section{Conclusion}

This paper presented a disentangled liquid latent dynamics model for
interpretable turbofan degradation modeling. The method encodes a history of
sensor readings and operating settings into a latent state, learns liquid
state increments \(\Delta z\) to roll that state forward, and decodes the
resulting trajectory into future sensor predictions. By separating degradation
and condition components and applying RUL, monotonicity, and latent-consistency
losses to the degradation state, the model encourages \(z^{\mathrm{deg}}\) to
behave as an inspectable health-state coordinate.

Experiments on C-MAPSS show that the full model improves sensor forecasting
overall, with the largest gains on the multi-condition FD002 and FD004 subsets.
The ablation study and PCA visualizations further show that the learned
degradation state exhibits an ordered trajectory, and that the magnitude of
latent state change aligns with degradation progress. At the same time, direct
RUL RMSE remains stronger for the GRU baseline, and the condition state still
contains some degradation leakage. The resulting conclusion is deliberately
bounded: liquid latent dynamics is promising for learning interpretable
degradation-state evolution and sensor world models, but further work is needed
to turn this representation into a fully calibrated lifetime predictor.

\bibliographystyle{plainnat}
\bibliography{references}

@inproceedings{saxena2008damage,
  title = {Damage Propagation Modeling for Aircraft Engine Run-to-Failure Simulation},
  author = {Saxena, Abhinav and Goebel, Kai and Simon, Don and Eklund, Neil},
  booktitle = {Proceedings of the International Conference on Prognostics and Health Management},
  year = {2008}
}

@inproceedings{zheng2017lstm,
  title = {Long Short-Term Memory Network for Remaining Useful Life Estimation},
  author = {Zheng, Shuai and Ristovski, Kosta and Farahat, Ahmed and Gupta, Chetan},
  booktitle = {Proceedings of the IEEE International Conference on Prognostics and Health Management},
  year = {2017}
}

@article{li2018dcnn,
  title = {Remaining Useful Life Estimation in Prognostics Using Deep Convolution Neural Networks},
  author = {Li, Xiang and Ding, Qian and Sun, Jian-Qiao},
  journal = {Reliability Engineering \& System Safety},
  volume = {172},
  pages = {1--11},
  year = {2018}
}

@article{gugulothu2017embed,
  title = {Predicting Remaining Useful Life Using Time Series Embeddings Based on Recurrent Neural Networks},
  author = {Gugulothu, Narendhar and TV, Vishnu and Malhotra, Pankaj and Vig, Lovekesh and Agarwal, Puneet and Shroff, Gautam},
  journal = {arXiv preprint arXiv:1709.01073},
  year = {2017}
}

@article{jayasinghe2019tcmn,
  title = {Temporal Convolutional Memory Networks for Remaining Useful Life Estimation of Industrial Machinery},
  author = {Jayasinghe, Lahiru and Samarasinghe, Tharaka and Yuen, Chau and Low, Jenny Chen Ni and Ge, Shuzhi Sam},
  journal = {arXiv preprint arXiv:1810.05644},
  year = {2019}
}

@inproceedings{ha2018worldmodels,
  title = {Recurrent World Models Facilitate Policy Evolution},
  author = {Ha, David and Schmidhuber, J{\"u}rgen},
  booktitle = {Advances in Neural Information Processing Systems},
  year = {2018}
}

@inproceedings{hafner2019planet,
  title = {Learning Latent Dynamics for Planning from Pixels},
  author = {Hafner, Danijar and Lillicrap, Timothy and Fischer, Ian and Villegas, Ruben and Ha, David and Lee, Honglak and Davidson, James},
  booktitle = {Proceedings of the International Conference on Machine Learning},
  year = {2019}
}

@inproceedings{chen2018neuralode,
  title = {Neural Ordinary Differential Equations},
  author = {Chen, Ricky T. Q. and Rubanova, Yulia and Bettencourt, Jesse and Duvenaud, David},
  booktitle = {Advances in Neural Information Processing Systems},
  year = {2018}
}

@inproceedings{rubanova2019latentode,
  title = {Latent Ordinary Differential Equations for Irregularly-Sampled Time Series},
  author = {Rubanova, Yulia and Chen, Ricky T. Q. and Duvenaud, David},
  booktitle = {Advances in Neural Information Processing Systems},
  year = {2019}
}

@inproceedings{hasani2021ltc,
  title = {Liquid Time-Constant Networks},
  author = {Hasani, Ramin and Lechner, Mathias and Amini, Alexander and Rus, Daniela and Grosu, Radu},
  booktitle = {Proceedings of the AAAI Conference on Artificial Intelligence},
  year = {2021}
}

@article{jardine2006review,
  title = {A Review on Machinery Diagnostics and Prognostics Implementing Condition-Based Maintenance},
  author = {Jardine, Andrew K. S. and Lin, Daming and Banjevic, Dragan},
  journal = {Mechanical Systems and Signal Processing},
  volume = {20},
  number = {7},
  pages = {1483--1510},
  year = {2006}
}

@article{si2011review,
  title = {Remaining Useful Life Estimation: A Review on the Statistical Data Driven Approaches},
  author = {Si, Xiao-Sheng and Wang, Wenbin and Hu, Chang-Hua and Zhou, Dong-Hua},
  journal = {European Journal of Operational Research},
  volume = {213},
  number = {1},
  pages = {1--14},
  year = {2011}
}

@article{lei2018review,
  title = {Machinery Health Prognostics: A Systematic Review from Data Acquisition to RUL Prediction},
  author = {Lei, Yaguo and Li, Naipeng and Guo, Liang and Li, Ningbo and Yan, Tao and Lin, Jing},
  journal = {Mechanical Systems and Signal Processing},
  volume = {104},
  pages = {799--834},
  year = {2018}
}

@inproceedings{heimes2008rnn,
  title = {Recurrent Neural Networks for Remaining Useful Life Estimation},
  author = {Heimes, Felix O.},
  booktitle = {Proceedings of the International Conference on Prognostics and Health Management},
  year = {2008}
}

@inproceedings{babu2016dcnn,
  title = {Deep Convolutional Neural Network Based Regression Approach for Estimation of Remaining Useful Life},
  author = {Babu, Giduthuri Sateesh and Zhao, Peilin and Li, Xiao-Li},
  booktitle = {Proceedings of the International Conference on Database Systems for Advanced Applications},
  pages = {214--228},
  year = {2016}
}

@inproceedings{malhotra2016lstm,
  title = {LSTM-Based Encoder-Decoder for Multi-Sensor Anomaly Detection},
  author = {Malhotra, Pankaj and Ramakrishnan, Anusha and Anand, Gaurangi and Vig, Lovekesh and Agarwal, Puneet and Shroff, Gautam},
  booktitle = {Proceedings of the ICML Workshop on Anomaly Detection},
  year = {2016}
}

@inproceedings{vaswani2017attention,
  title = {Attention Is All You Need},
  author = {Vaswani, Ashish and Shazeer, Noam and Parmar, Niki and Uszkoreit, Jakob and Jones, Llion and Gomez, Aidan N. and Kaiser, Lukasz and Polosukhin, Illia},
  booktitle = {Advances in Neural Information Processing Systems},
  year = {2017}
}

@article{lim2021tft,
  title = {Temporal Fusion Transformers for Interpretable Multi-Horizon Time Series Forecasting},
  author = {Lim, Bryan and Arik, Sercan O. and Loeff, Nicolas and Pfister, Tomas},
  journal = {International Journal of Forecasting},
  volume = {37},
  number = {4},
  pages = {1748--1764},
  year = {2021}
}

@inproceedings{zhou2021informer,
  title = {Informer: Beyond Efficient Transformer for Long Sequence Time-Series Forecasting},
  author = {Zhou, Haoyi and Zhang, Shanghang and Peng, Jieqi and Zhang, Shuai and Li, Jianxin and Xiong, Hui and Zhang, Wancai},
  booktitle = {Proceedings of the AAAI Conference on Artificial Intelligence},
  year = {2021}
}

@inproceedings{chung2015vrnn,
  title = {A Recurrent Latent Variable Model for Sequential Data},
  author = {Chung, Junyoung and Kastner, Kyle and Dinh, Laurent and Goel, Kratarth and Courville, Aaron and Bengio, Yoshua},
  booktitle = {Advances in Neural Information Processing Systems},
  year = {2015}
}

@article{krishnan2015dvbf,
  title = {Deep Kalman Filters},
  author = {Krishnan, Rahul G. and Shalit, Uri and Sontag, David},
  journal = {arXiv preprint arXiv:1511.05121},
  year = {2015}
}

@inproceedings{fraccaro2016dkf,
  title = {Sequential Neural Models with Stochastic Layers},
  author = {Fraccaro, Marco and S{\o}nderby, S{\o}ren Kaae and Paquet, Ulrich and Winther, Ole},
  booktitle = {Advances in Neural Information Processing Systems},
  year = {2016}
}

@article{lechner2020ncp,
  title = {Neural Circuit Policies Enabling Auditable Autonomy},
  author = {Lechner, Mathias and Hasani, Ramin and Amini, Alexander and Henzinger, Thomas A. and Rus, Daniela and Grosu, Radu},
  journal = {Nature Machine Intelligence},
  volume = {2},
  pages = {642--652},
  year = {2020}
}

@article{hasani2022cfc,
  title = {Closed-Form Continuous-Time Neural Networks},
  author = {Hasani, Ramin and Lechner, Mathias and Amini, Alexander and Rus, Daniela and Grosu, Radu},
  journal = {Nature Machine Intelligence},
  volume = {4},
  pages = {992--1003},
  year = {2022}
}

@article{bengio2013representation,
  title = {Representation Learning: A Review and New Perspectives},
  author = {Bengio, Yoshua and Courville, Aaron and Vincent, Pascal},
  journal = {IEEE Transactions on Pattern Analysis and Machine Intelligence},
  volume = {35},
  number = {8},
  pages = {1798--1828},
  year = {2013}
}

@inproceedings{higgins2017betavae,
  title = {beta-VAE: Learning Basic Visual Concepts with a Constrained Variational Framework},
  author = {Higgins, Irina and Matthey, Loic and Pal, Arka and Burgess, Christopher and Glorot, Xavier and Botvinick, Matthew and Mohamed, Shakir and Lerchner, Alexander},
  booktitle = {Proceedings of the International Conference on Learning Representations},
  year = {2017}
}

@inproceedings{locatello2019challenging,
  title = {Challenging Common Assumptions in the Unsupervised Learning of Disentangled Representations},
  author = {Locatello, Francesco and Bauer, Stefan and Lucic, Mario and Raetsch, Gunnar and Gelly, Sylvain and Schoelkopf, Bernhard and Bachem, Olivier},
  booktitle = {Proceedings of the International Conference on Machine Learning},
  year = {2019}
}

@inproceedings{zhao2017health,
  title = {Deep Learning and Its Applications to Machine Health Monitoring},
  author = {Zhao, Rui and Yan, Ruqiang and Chen, Zhenghua and Mao, Kezhi and Wang, Peng and Gao, Robert X.},
  booktitle = {Mechanical Systems and Signal Processing},
  year = {2019}
}

@article{schwabacher2005survey,
  title = {A Survey of Data-Driven Prognostics},
  author = {Schwabacher, Mark},
  journal = {AIAA Infotech@Aerospace Conference},
  year = {2005}
}

@article{gao2020survey,
  title = {A Survey of Deep Learning for Remaining Useful Life Prediction},
  author = {Gao, Yifan and Zhu, Yuxuan and Li, Xin and Wang, Yaguo},
  journal = {arXiv preprint arXiv:2006.06414},
  year = {2020}
}

@article{su2019pooled,
  title = {Pooled Time Series Representation for Mitosis Event Recognition},
  author = {Su, Yuting and Wang, Shan and Nie, Weizhi and An, Yang},
  journal = {Multimedia Systems},
  year = {2019}
}

@article{nie2018mitosis,
  title = {Mitosis Event Recognition and Detection Based on Evolution of Feature in Time Domain},
  author = {Nie, Weizhi and Yan, Yan and Hao, Tong and Liu, Chenchen and Su, Yuting},
  journal = {Machine Vision and Applications},
  volume = {29},
  number = {8},
  pages = {1249--1256},
  year = {2018}
}

@article{nie2023icu,
  title = {Temporal-Spatial Correlation Attention Network for Clinical Data Analysis in Intensive Care Unit},
  author = {Nie, Weizhi and Yu, Yuhe and Zhang, Chen and Song, Dan and Zhao, Lina and Bai, Yunpeng},
  journal = {IEEE Transactions on Biomedical Engineering},
  year = {2023}
}

@article{li2023brain,
  title = {Brain Tumor Image Segmentation Based on Prior Knowledge via Transformer},
  author = {Li, Qiang and Liu, Hengxin and Nie, Weizhi and Wu, Ting},
  journal = {International Journal of Imaging Systems and Technology},
  year = {2023}
}

@article{nie2024t2td,
  title = {T2TD: Text-3D Generation Model Based on Prior Knowledge Guidance},
  author = {Nie, Weizhi and Chen, Ruidong and Wang, Weijie and Lepri, Bruno and Sebe, Nicu},
  journal = {IEEE Transactions on Pattern Analysis and Machine Intelligence},
  pages = {1--18},
  year = {2024}
}

@article{li2024cisepsis,
  title = {CISepsis: A Causal Inference Framework for Early Sepsis Detection},
  author = {Li, Qiang and Li, Dongchen and Jiao, He and Wu, Zhenhua and Nie, Weizhi},
  journal = {Frontiers in Cellular and Infection Microbiology},
  year = {2024}
}

@article{zhang2025cabg,
  title = {Causal Inference Model for Accurate Medical Diagnosis in Coronary Artery Bypass Graft Operation},
  author = {Zhang, Qiyi and Zhang, Wei and Li, Qiang and Bai, Yunpeng and Nie, Weizhi and Xie, Keliang},
  journal = {Artificial Intelligence in Medicine},
  year = {2025}
}

\end{document}